\documentclass{article}
\usepackage{ijcai21}

\usepackage{times}

\usepackage{soul}
\usepackage{color}
\usepackage{url}
\usepackage[hidelinks]{hyperref}
\usepackage[utf8]{inputenc}
\usepackage[small]{caption}
\usepackage{graphicx}
\usepackage{amsmath}
\usepackage{booktabs}
\newcommand*\samethanks[1][\value{footnote}]{\footnotemark[#1]}

\title{Parsing Birdsong with Deep Audio Embeddings}
\author{
Irina Tolkova$^1$\thanks{equal contribution}\and
Brian Chu$^1$\samethanks\and
Marcel Hedman$^1$\samethanks\and
Stefan Kahl$^2$\And
Holger Klinck$^2$\\
\affiliations
$^1$School of Engineering and Applied Sciences, Harvard University, Cambridge, MA\\
$^2$K. Lisa Yang Center for Conservation Bioacoustics, Cornell Lab of Ornithology, Cornell University, Ithaca, NY\\
\emails
\{itolkova, brianchu, marcelhedman\}@g.harvard.edu,
sk2487@cornell.edu,
holger.klinck@cornell.edu
}

\begin{document}

\maketitle

\begin{abstract}
Monitoring of bird populations has played a vital role in conservation efforts and in understanding biodiversity loss. The automation of this process has been facilitated by both sensing technologies, such as passive acoustic monitoring, and accompanying analytical tools, such as deep learning. However, machine learning models frequently have difficulty generalizing to examples not encountered in the training data. In our work, we present a semi-supervised approach to identify characteristic calls and environmental noise. We utilize several methods to learn a latent representation of audio samples, including a convolutional autoencoder and two pre-trained networks, and group the resulting embeddings for a domain expert to identify cluster labels. We show that our approach can improve classification precision and provide insight into the latent structure of environmental acoustic datasets.
\end{abstract}

\section{Introduction}

Over the past half-century, there has a been an estimated decline of 29\% across North American bird populations, corresponding to a net loss of about 3 billion birds \cite{rosenberg2019decline}. To design and implement effective conservation policies for counteracting such biodiversity loss, there is a need for automated monitoring technologies to understand changes in species-level populations. In recent years, technologies such as camera traps \cite{rowcliffe2008surveys,steenweg2017scaling,o2010camera}, drone footage \cite{jimenez2019drones,van2014nature,koh2012dawn}, or even satellite imagery \cite{pettorelli2014satellite} have been used for species-level identification and population estimation. However, these visual methods are difficult to apply to bird recognition, due to their small size and frequent occlusion in the environment. But a majority of birds are vocal, with complex species-specific calls and song repertoires. Consequently, there has been a rising interest in using automated passive acoustic technologies for monitoring bird populations \cite{browning2017passive,sugai2019terrestrial,blumstein2011acoustic,wood2019}.

In particular, BirdNET is an acoustic monitoring framework drawing on citizen science for large-scale data collection on species-level occurrence \cite{kahl2021birdnet}. The foundation of BirdNET is a CNN classifier, trained on high-quality recordings from labeled birdsong repositories such as Xeno-Canto (www.xeno-canto.org) and the Macaulay Library at the Cornell Lab of Ornithology (https://www.macaulaylibrary.org). A user can record birdsong through the BirdNET app and see the species identified by the classifier, across three thousand possible species. Moreover, these audio recordings, classification outputs, and associated time/location metadata are anonymized and catalogued. Therefore, the project intends to both engage the public in bird-watching while also collecting large-scale data for spatiotemporal analysis of avian populations.

While the use of machine learning mitigates the need for expert labeling and opens opportunities in data collection, it is imperfect, and will have difficulties generalizing to sound structure not encountered in the training dataset. This is of particular significance for BirdNET, which has a domain shift between the training and deployment datasets. Xeno-Canto and the Macaulay Library contain recordings with high signal-to-noise ratio obtained with high-quality (often directional) microphones, from which it is easy to obtain labels localized in time (“strong” labels). However, user submissions are often taken with a cell phone in noisy environments, potentially within a long recording window. To improve robustness, the training pipeline incorporates data augmentation with various sources of “background noise” data; but submissions will likely contain noise structure which is not encompassed by these augmentations. Overall, evaluations of BirdNET have shown false positive rates around 15-20\% \cite{arif2020testing}. While this accuracy outperforms other classifiers, and may be sufficient for citizen-science engagement, such a margin is significant in informing conservation policy.

In this work, we aim to introduce a semi-supervised approach for post-processing BirdNET classifications and identifying false positives. Specifically, we utilize several methods for representation learning to construct embeddings of acoustic data and identify clusters characteristic of calls or environmental noise. By incorporating human input on the content of each cluster, we provide alternative confidence scores to the BirdNET classifications. We evaluate this approach on a collection of recordings across three species which were all identified as ``positives" by the BirdNET system and were subsequently manually evaluated by an expert, with the primary goal of improving system precision. We hope that this project can improve the performance of automated passive acoustic monitoring, inform the targeted system augmentations needed to address false positives, and provide insights into lower-dimensional structure within intraspecific acoustic data.

\section{Related Work}

The introduction of AudioSet has enabled many recent advances in audio event recognition. AudioSet is a collection of short human-labeled sound clips from YouTube videos, containing over 2 million samples separated into 632 audio classes by human annotation \cite{gemmeke2017audioset}. Most deep learning approaches for audio analysis will first generate spectrograms from the data, turning the task into an image classification problem. VGGish is a pre-trained convolutional neural network inspired by the VGG network used for image classification \cite{hershey2017cnn}. Other similar approaches include SoundNet \cite{Aytar2016SoundNetLS} and $L^3$-net \cite{arandjelovic2017look,arandjelovic2018objects,cramer2019look} which learn representations on unlabelled data. More recently, studies have leveraged triplet learning or contrastive learning to generate audio embeddings. These approaches are able to use the raw audio signal \cite{yu2020contrastive} or its spectrogram \cite{Spijkervet2021ContrastiveLO}.

%These techniques use unsupervised learning to cluster similar audio samples and separate dissimilar sounds.

While there have been major advances in audio embeddings within machine hearing, they have had limited use in passive acoustic monitoring. Unsupervised methods such as clustering techniques, t-distributed Stochastic Neighbor Embedding (t-SNE) and Uniform Manifold Approximation (UMAP) have previously been applied for distinguishing calls \cite{clink2021unsupervised,parra2020uniform}. In particular, \cite{sainburg2020finding} performed a thorough analysis of UMAP applied to clustering across intraspecific vocalizations in a range of species. Of studies that have leveraged deep learning, \cite{sethi2020characterizing} used VGGish, pre-trained on AudioSet, to analyze soundscape data. The authors found that the embedding captured clear structure across spatio-temporal characteristics, and demonstrated the use of latent distributions for acoustic anomaly detection.

\section{Methods}

\subsection{Pre-processing and Spectrogram Calculation}

% NOTE: these processing steps may change!! But putting them here for the time being.

We consider BirdNET submissions from 3 species -- the barred owl ({\it{Strix varia}}), common crane ({\it{Grus grus}}), and common loon ({\it{Gavia immer}}) -- chosen for common occurrence, diversity in vocalization characteristics, and high false positive rates. Each recording is subdivided into windows of length 1 sec, which overlap by 0.5 sec to ensure better coverage of acoustic features (for instance, a recording of length 3 seconds would yield 5 windows). We then calculate spectrograms with a Hann taper function of size 1024, and a shift of 25\%. We normalize spectrograms by maximum amplitude, convert amplitudes to log-scale (with a regularization term of 0.001), and convert frequency to the mel-scale. As most bird calls occur below 10kHz, we truncate the frequency range to this value. To remove low-amplitude windows below a baseline level of acoustic activity, we apply a simple detector score similar to that used in \cite{kahl2021birdnet,sprengel2016audio}. For a given spectrogram, the detector score is defined by the proportion of pixels that are greater than the column median and 1.5 times the row median. For the BirdNET dataset, we ignore all windows with a score of less than 0.1; on the other hand, for obtaining ``call" samples from the Xeno-Canto dataset, we use only windows scoring above 0.3. These thresholds were determined by visual examination. Lastly, we downsize the windows to a size of 100x100 pixels for computational feasibility. These spectrograms are used as input for the autoencoder, and for visualization across all methods; the corresponding time-domain signal is given as input to both pre-trained architectures (VGGish and PANNs).

\subsection{Embeddings}

\subsubsection{Pre-trained VGGish}
The first embedding method explored uses the VGGish architecture pretrained on AudioSet, following the methodology described by \cite{sethi2020characterizing}. This approach has been shown to successfully identify anomalous sounds in a natural environment, but has not been applied to the more specific problem of identifying intraspecific calls. After following the pre-processing steps outlined above, 1-second time-domain samples are given as input into the pre-trained architecture, which converts the signal into a spectrogram and returns a 128-dimensional embedding.

\subsubsection{Pre-trained Wavegram-Logmel-CNN}

While most audio analysis is performed either on spectrogram images or (more rarely) on raw audio, \cite{kong2020panns} proposes integrating these approaches for the Wavegram-Logmel-CNN16 architecture: using both a learned 1-dimensional time-frequency representation together with the calculated log-amplitude mel-scale spectrogram as input. This network, pretrained on AudioSet, gives superior performance to prior methods on a number of downstream classification tasks with a 2048-dimensional embedding, and is made publicly available with a user-friendly interface through the {\it{panns-inference}} Python package.

\subsubsection{Autoencoder}

We also obtained embeddings through the use of an convolutional autoencoder: a network which attempts to learn a representation of the original data, typically in a lower dimensionality. An autoencoder is trained to minimize the reconstruction loss, or difference between the input image (in this case, the mel-spectrogram) and the output image. The bottleneck layer serves to capture the most important signal from the input, and we use this layer to extract the embeddings. We chose to use 10 layers and an embedding size of 128; the specific architecture can be found in Table \ref{tab:ae}.

\begin{table}[h]
\centering
\begin{tabular}{c||c|c|c|c}
\toprule
Layer & Operation                                                           & In Size                                                               & Out Size                                                         & Kernel  \\ \midrule
1     & \begin{tabular}[c]{@{}c@{}}conv\\ relu\end{tabular}                 & \begin{tabular}[c]{@{}c@{}}100x100x1\\ 49x49x32\end{tabular}          & \begin{tabular}[c]{@{}c@{}}49x49x32\\ 49x49x32\end{tabular}      & 4x4          \\ \hline
2     & \begin{tabular}[c]{@{}c@{}}conv\\ relu\end{tabular}                 & \begin{tabular}[c]{@{}c@{}}49x49x32\\ 23x23x64\end{tabular}           & \begin{tabular}[c]{@{}c@{}}23x23x64\\ 23x23x64\end{tabular}      & 4x4        \\ \hline
3     & \begin{tabular}[c]{@{}c@{}}conv\\ relu\end{tabular}                 & \begin{tabular}[c]{@{}c@{}}23x23x64\\ 10x10x128\end{tabular}          & \begin{tabular}[c]{@{}c@{}}10x10x128\\ 10x10x128\end{tabular}    & 4x4          \\ \hline
4     & \begin{tabular}[c]{@{}c@{}}conv\\ relu\\ flatten\end{tabular}       & \begin{tabular}[c]{@{}c@{}}10x10x128\\ 4x4x256\\ 4x4x256\end{tabular} & \begin{tabular}[c]{@{}c@{}}4x4x256\\ 4x4x246\\ 4096\end{tabular} & 4x4          \\ \hline
5     & fc                                                     & 4096                                                                  & 128                                                              &                \\ \hline
6     & \begin{tabular}[c]{@{}c@{}}fc\\ unflatten\end{tabular} & \begin{tabular}[c]{@{}c@{}}128\\ 4096\end{tabular}                    & \begin{tabular}[c]{@{}c@{}}4096\\ 1x1x4096\end{tabular}                 &        \\ \hline
7     & \begin{tabular}[c]{@{}c@{}}conv\\ relu\end{tabular}                 & \begin{tabular}[c]{@{}c@{}}1x1x4096\\ 7x7x128\end{tabular}            & \begin{tabular}[c]{@{}c@{}}7x7x128\\ 7x7x128\end{tabular}        & 7x7          \\ \hline
8     & \begin{tabular}[c]{@{}c@{}}conv\\ relu\end{tabular}                 & \begin{tabular}[c]{@{}c@{}}7x7x128\\ 20x20x64\end{tabular}            & \begin{tabular}[c]{@{}c@{}}20x20x64\\ 20x20x64\end{tabular}      & 8x8         \\ \hline
9     & \begin{tabular}[c]{@{}c@{}}conv\\ relu\end{tabular}                 & \begin{tabular}[c]{@{}c@{}}20x20x64\\ 47x47x32\end{tabular}           & \begin{tabular}[c]{@{}c@{}}47x47x32\\ 47x47x32\end{tabular}      & 9x9          \\ \hline
10    & \begin{tabular}[c]{@{}c@{}}conv\\ sigmoid\end{tabular}              & \begin{tabular}[c]{@{}c@{}}47x47x32\\ 100x100x1\end{tabular}          & \begin{tabular}[c]{@{}c@{}}100x100x1\\ 100x100x1\end{tabular}    & 8x8        
\end{tabular}
\caption{Autoencoder architecture. A stride of 2 was used for all layers.}
\label{tab:ae}
\end{table}

\begin{figure}
\includegraphics[width=8cm]{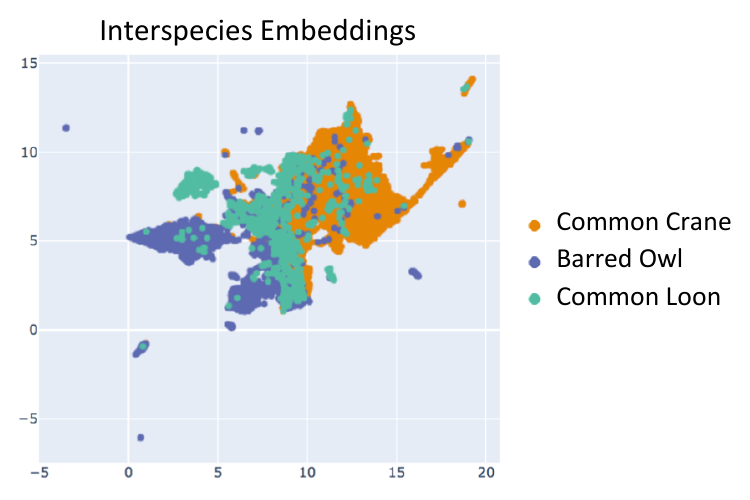}
\caption{\label{fig:intraspecific}{This figure shows the Wavegram-Logmel-CNN16 embeddings for the Xeno-Canto samples for each species, projected to 2 dimensions with UMAP.}}
\end{figure}

\subsection{Classification and Evaluation}

To group similar samples, we apply k-means and extract clusters in the high-dimensional embedding space. While it is possible to automatically select the cluster number using the silhouette coefficient, we opt to fix this value to 12 for consistency across species, as the disadvantage of a higher number of clusters is human labeling effort rather than a poor fit to data. Next, we randomly sample a collection of spectrograms from each cluster to present to an expert, and retrieve a binary label indicating whether a cluster generally represents bird calls. Since a species may have a varied repertoire of calls, we may expect multiple disconnected clusters to be assigned a positive label. To visualize and evaluate clustering quality and latent structure, we used UMAP to project the high-dimensional embeddings to two dimensions.

We then infer cluster labels (and consequently associated positive/negative labels) for the BirdNET audio. Specifically, for each BirdNET sample, we locate Xeno-Canto samples within a pre-defined threshold distance, and assign the sample to the most commonly occurring cluster among these neighbors. Finally, to obtain labels for recordings which can be compared against ground-truth labels, we give a positive label to a BirdNET submission if it contains at least two positively-labeled samples. We then evaluate accuracy, with a particular interest in precision, as false positives are considered more detrimental to ecological study of population dynamics than false negatives. Moreover, since all of the BirdNET recordings used were positive classifications from the BirdNET system, we consider the {\it{improvement in precision}} through the integration of this semi-supervised pipeline.

\begin{figure*}
\includegraphics[width=7in]{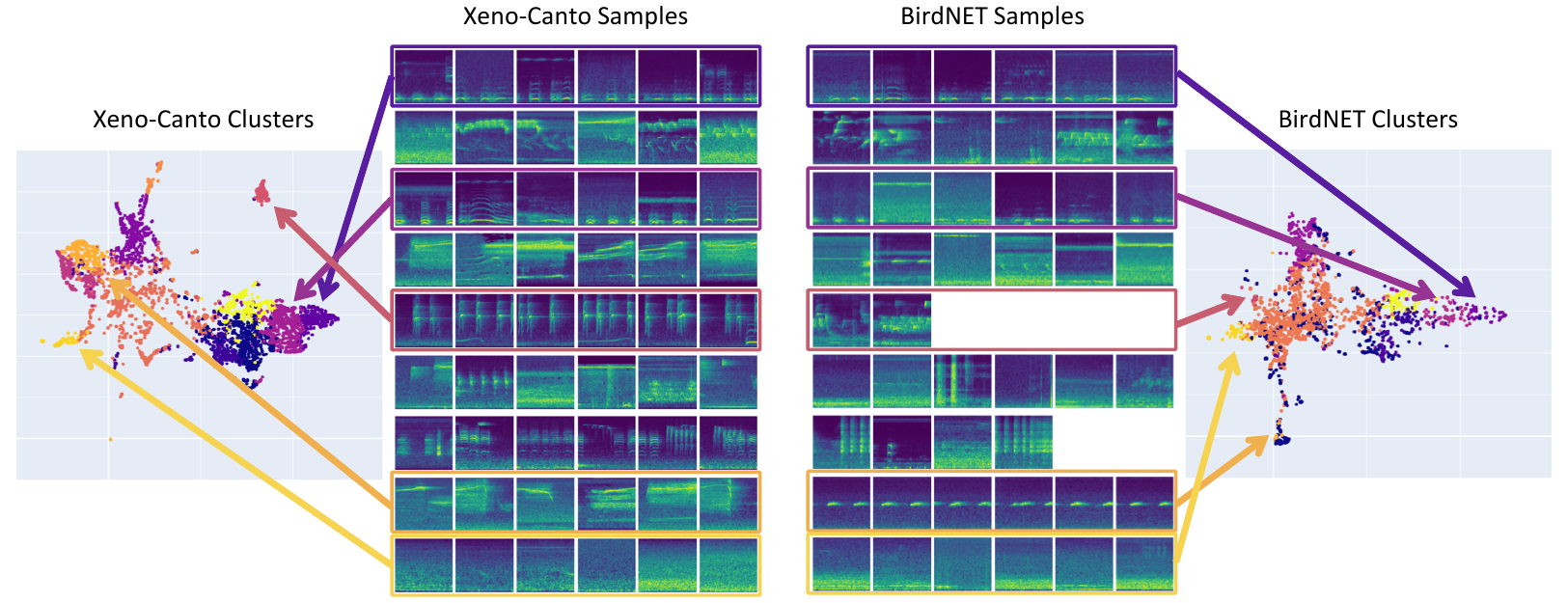}
\caption{\label{fig:clusters}{This figure shows clusters in the embedding space of the Wavegram-Logmel-CNN16 network, together with spectrogram samples corresponding to a subset of the clusters, for both Xeno-Canto (left half) and BirdNET (right half).}}
\end{figure*}

\begin{table*}[h]
\centering
\begin{tabular}{l|c|c|c|c|c|c|c|c|c}
                    & \multicolumn{3}{c|}{\textbf{Barred Owl}} & \multicolumn{3}{c|}{\textbf{Common Crane}} & \multicolumn{3}{c}{\textbf{Common Loon}} \\ \hline
\textbf{Architecture}        & Accuracy  & Precision  & Recall & Accuracy   & Precision  & Recall  & Accuracy  & Precision  & Recall  \\ \hline
VGG-ish             &      0.65     & 0.52 (\textbf{+.15})        & 0.61    &       0.47     &   0.40 (-.03)     &   0.45      &     0.58      &       0.05 (\textbf{+.01})     &     0.44    \\ \hline

W-L-CNN16 & 0.73          & 0.74 (\textbf{+.37})     & 0.43     &           0.55 & 0.48   (\textbf{+.05})        &   0.36      & 0.95          &  0.33    (\textbf{+.29})      &    0.22     \\ \hline
Autoencoder         &    0.57       & 0.44 (\textbf{+.07})       & 0.63    &   0.40         &      0.39   (-.04)   &     0.71    &      0.96     &     0.50  (\textbf{+.46})    & 0.11
         \\ 
\end{tabular}
\caption{Classification results for each audio embedding method and the three species. The value in parentheses is the improvement over BirdNET precision.}
\label{tab:results}
\end{table*}

% \begin{figure}
% \includegraphics[width=8cm]{images/Carolina Wren VGGish call vs non call.png}
% \caption{\label{fig:FIG1}{Figure showing call vs non-call in the embedding space generated by the VGGish architecture. Species: Carolina Wren. Yellow: call, Blue:no call}}
% \end{figure}

\section{Results and Discussion}

First, we compare the UMAP projections of embeddings across species with the different methods (Figure \ref{fig:intraspecific}). Each point represents a 1-second audio sample from the Xeno-Canto recordings. Note that we expect overlap in these distributions, as a significant part of the dataset will contain background noise rather than species-specific vocalizations; but we also see separation of clusters across the three species.

Next, we continue to the more complex task of intraspecific analysis. We visually examine the projected Xeno-Canto and BirdNET distributions, and find that all methods produce meaningful latent structure and clustering. For example, Figure \ref{fig:clusters} demonstrates the projected embedding obtained through the Wavegram-Logmel-CNN16 for the barred owl dataset. The Xeno-Canto embedding colored by cluster assignment is on the far left, and randomly-selected samples from nine of these clusters are shown in the spectrograms on the left. The associated BirdNET embedding, colored by inferred cluster label, along with corresponding spectrogram samples, is on the right. We see that the clusters are directly interpretable: the first and third rows capture mature barred owl hoots; the fourth and eighth rows are primarily begging calls by barred owl chicks; the last row describes wide-band environmental-noise; and the second, fifth, and seventh represent vocalizations from other species which unintentionally occur in the dataset. By providing human-in-the-loop input on such erroneous training samples, this pipeline could help to remove them from analysis: whether prior to training, or in post-processing as described here. Additionally, the locations of these clusters in the embedding space can also be interpreted: clusters representing hoots are close together, and far from the two clusters representing juvenile begging calls, with points representing other noises scattered in between. By comparing the two embedding distributions, we can interpret the domain shift in the data: some of the clusters result in similar samples between Xeno-Canto and BirdNET, while others (such as clusters 5 and 8) change significantly. Moreover, we see that a greater proportion of the BirdNET samples are found in the "noisy" clusters, rather than in the "hoot" clusters on the right edge of the plot: this result is expected, since we impose lower pre-processing detection thresholds on BirdNET submissions to avoid omission of quiet calls.

%Additionally, by superimposing the two datasets, we can visualize the domain shift occurring in the data. Specifically, we see that the BirdNET submissions are more heavily distributed in regions containing the ``background noise" clusters. This result is expected, since BirdNET submissions are typically longer in length, containing no bird sounds towards the beginning and end of recordings.

Next, Table \ref{tab:results} displays the accuracies, precision, and recall across the three species for each of the three methods. On the whole, we find that the methods can assist in distinguishing recordings containing true positive samples from the false positives, but further development is necessary to improve accuracy and robustness. In general, the Wavegram-Logmel-CNN16 outperforms the two other techniques in precision, though has lower recall. We see best performance for the barred owl, in which we can obtain a significant precision improvement. However, the low number of positive samples within this data caused significant difficulty for classification. For all three species, fewer than half of the recordings were correct -- in particular, in the available data for the common loon, only 9 of the 210 recordings contained true identification of this species by BirdNET. Consequently, misclassifications of several spectrograms within our pipeline would cause significant drops in precision and accuracy.

%even increasing precision for the Common Crane when the others did not. 

\section{Broader Impact}

Since the spatial distribution of both BirdNET submissions and Xeno-Canto recordings is not uniform across the globe, an analytical pipeline developed on this data may face geographically variable performance. Currently, most BirdNET submissions are concentrated in the US and Western European countries, as these regions received earlier deployment of the BirdNET app. In addition, while BirdNET is available internationally and supports a translated user interface across 11 languages, access to the app will be limited by smartphone prevalence and public awareness. Xeno-Canto has a greater geographic reach, with high representation in South America and Europe, and growing coverage across Asia and Africa. As different environments will have different ambient soundscapes, calls, and sources of noise, there should be discretion in extrapolating classification accuracy to new regions -- particularly since over-estimation of species population sizes may inhibit conservation measures for those species.

\section{Conclusion}

In this paper, we provide a framework for learning representations of BirdNET classifications and identifying false positives. Our approach can enable experts to examine the structural differences between bird calls and other audio samples, both on an inter- and intra-species level, and thereby detect misclassifications. Future work includes expanding this analysis to a wider collection of taxonomic groups (particularly to the passerines, or songbirds), and incorporating the false positive identification into the BirdNET classifier.

\section*{Acknowledgments}

We are very grateful to Daniel Salisbury for manually labeling the BirdNET data used in this work. We would like to thank Connor Wood and Ben Mirin for feedback and species identification. Additionally, we would like to thank Milind Tambe, Doria Spiegel, and Boriana Gjura for their support on this project.

\appendix

\bibliographystyle{named}
\bibliography{ijcai21}

\end{document}